\documentclass[11pt,a4paper]{article}
\usepackage[hyperref]{acl2021}
\usepackage{times}
\usepackage{latexsym}

\usepackage{microtype}
\usepackage{graphicx}
\usepackage{booktabs}  
\usepackage{multirow}
\usepackage{subfigure}
\usepackage{verbatim}
\usepackage{CJKutf8} 
\usepackage{flushend}
\usepackage{makecell}
\usepackage{amsmath}
\usepackage{amssymb}

\aclfinalcopy 

\title{Smart Bird: Learnable Sparse Attention for \\Efficient and Effective Transformer}

\author{Chuhan Wu$^1$~~Fangzhao Wu$^2$~~Tao Qi$^1$~~Binxing Jiao$^3$~~Daxin Jiang$^3$~~Yongfeng Huang$^1$~~Xing Xie$^2$\\
    $^1$Department of Electronic Engineering \& BNRist, Tsinghua University, Beijing 100084, China  \\
     $^2$Microsoft Research Asia, Beijing 100080, China\\
     $^3$Microsoft STC Asia, Beijing 100080, China\\
  {\tt\{wuchuhan15, wufangzhao, taoqi.qt\}@gmail.com} \\
  {\tt \{binxjia, djiang, xingx\}@microsoft.com}\\
  {\tt yfhuang@tsinghua.edu.cn}
  }
\date{}

\begin{document}
\maketitle

\begin{abstract}

Transformer has achieved great success in NLP.
However, the quadratic complexity of the self-attention mechanism in Transformer makes it inefficient in handling long sequences.
Many existing works explore to accelerate Transformers by computing sparse self-attention  instead of a dense one, which usually attends to tokens at certain positions or randomly selected tokens.
However, manually selected or random tokens may be uninformative for context modeling.
In this paper, we propose \textit{Smart Bird}, which is an efficient and effective Transformer with learnable  sparse attention.
In \textit{Smart Bird}, we first compute a sketched attention matrix with a single-head low-dimensional Transformer, which aims to find potential important interactions between tokens.
We then sample token pairs based on their probability scores derived from the sketched attention matrix to generate different sparse attention index matrices for different attention heads.
Finally, we select token embeddings according to the index matrices to form the input of sparse attention networks.
Extensive experiments on six benchmark datasets for different tasks validate the efficiency and effectiveness of \textit{Smart Bird}  in  text modeling.

\end{abstract}

\section{Introduction}

Transformer~\cite{vaswani2017attention} has achieved great success in NLP by serving as the basic architecture of many popular  models such as BERT~\cite{devlin2019bert}, RoBERTa~\cite{radford2019language} and XLM~\cite{conneau2019cross}.
In addition, Transformers also show great potentials in other fields like computer vision~\cite{dosovitskiy2020image} and speech recognition~\cite{dong2018speech} to understand data in other modalities.
Self-attention is the core of Transformer~\cite{parikh2016decomposable}.
It aims to model the interaction between each pair of tokens to help capture their contexts, where the computational complexity is quadratic with respect to the input sequence length~\cite{vaswani2017attention}.
Thus, Transformer is inefficient in handling long sequences~\cite{tay2020efficient}.

An intuitive way to improve the efficiency of Transformer is computing a sparse self-attention matrix to reduce the number of tokens to be attended~\cite{child2019generating}.
For example, \citet{beltagy2020longformer} proposed Longformer, which computes sliding window attention to capture local contexts and global attention at a few positions to capture global contexts.
\citet{zaheer2020big} proposed BidBird, which further incorporates random attention that models the interactions between each token with a certain number of randomly selected tokens.
However, attending to the tokens that are   randomly sampled or selected by certain rules may not be actually helpful for context modeling.

\begin{figure}[!t]
  \centering
  \subfigure[Attention heatmap of a 4-dim Transformer.]{
    \includegraphics[width=0.22\textwidth]{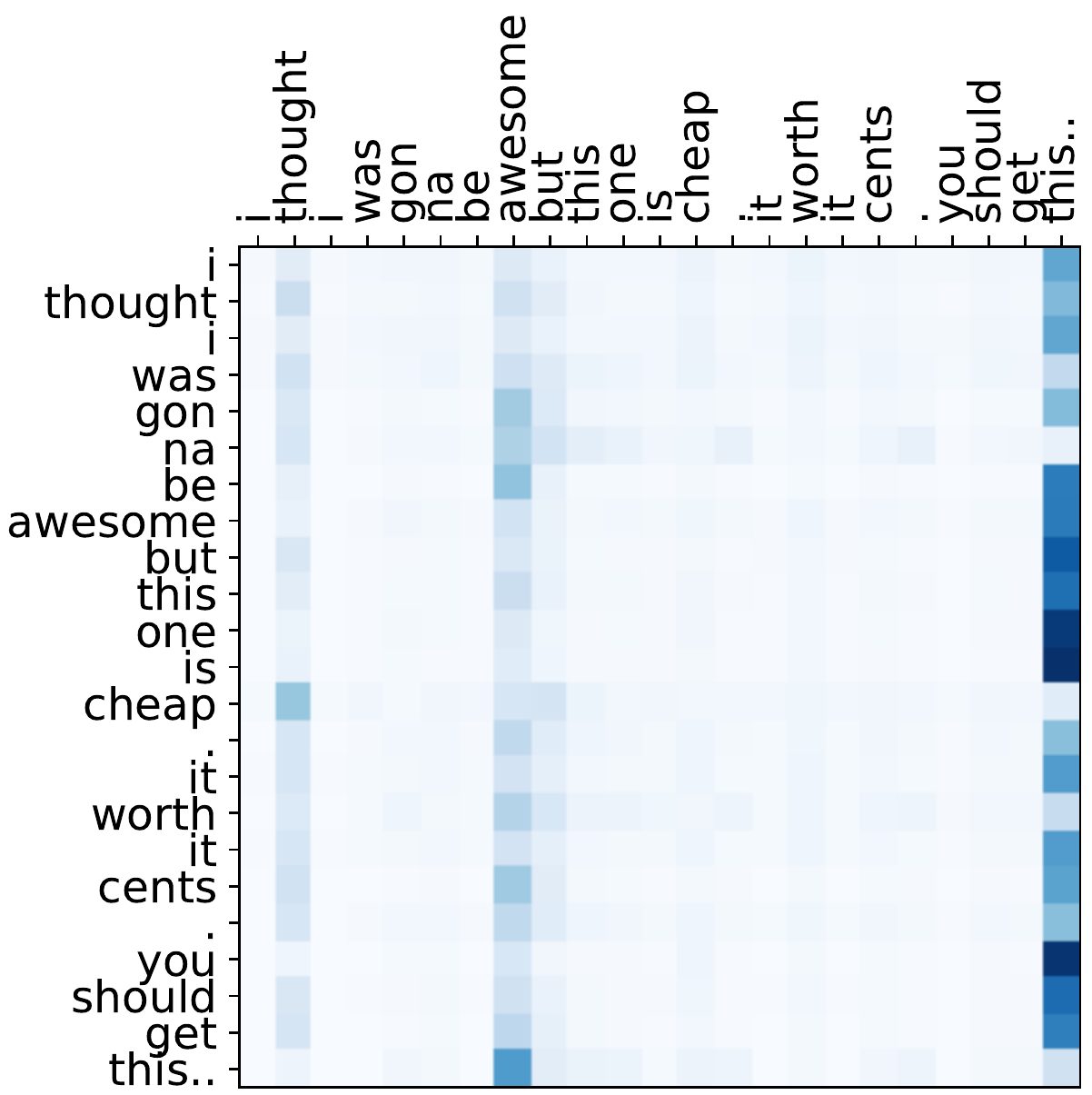}
  \label{fig.smallatt}
  }
   \subfigure[Attention heatmap of a 256-dim Transformer.]{
      \includegraphics[width=0.22\textwidth]{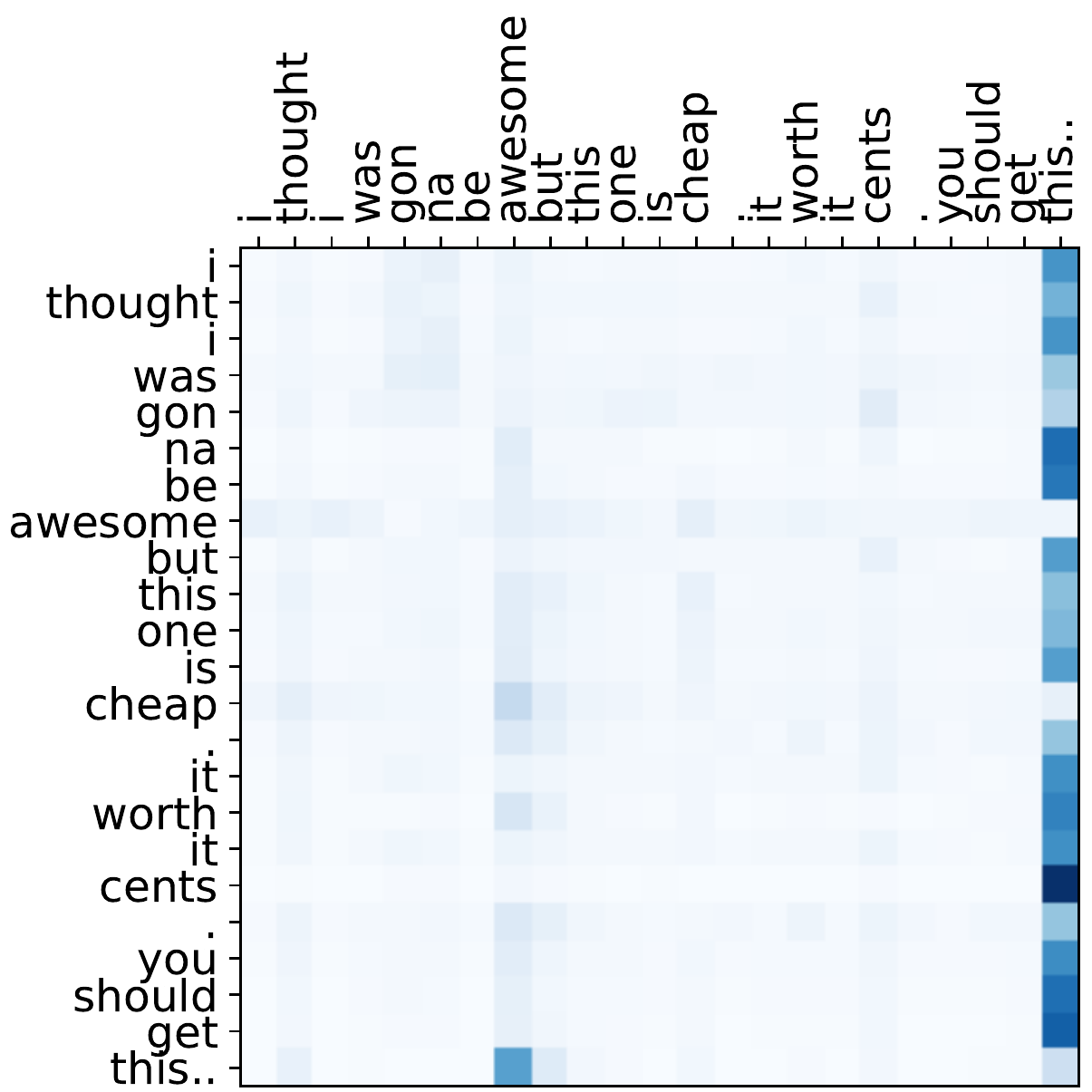}
  \label{fig.bigatt}
  }
  \caption{The attention heatmaps learned by Transformers with 4 or 256 hidden dimensions.}\label{fig.att}

\end{figure}

Instead of attending to the heuristically or randomly selected tokens, attending to those potentially important tokens may help build higher-quality sparse attention for text modeling.
Fortunately, these potentially important tokens can be efficiently and effectively identified by a tiny Transformer model with very low dimensions.
Fig.~\ref{fig.att} show the attention heatmaps learned by a tiny Transformer with 4 hidden dimensions and a standard 256-dim Transformer.
We find the attention heatmap produced by the low-dimensional Transformer is very similar to the heatmap learned by a Transformer with a much larger size.
In fact, on the Amazon dataset~\cite{he2016ups}, the average Pearson correlation coefficient between the attention matrices learned by Transformers with 256-dim or 4-dim attention heads is 0.67, while the correlation coefficient between attention matrices learned by the same Transformer in two independent experiments with different random seeds is  0.76. 
Thus, the attention heatmaps learned by the low-dimensional Transformer have the potential to guide the sparse attention mechanism by providing useful clues on recognizing important interactions among tokens to empower context modeling.

In this paper, we propose \textit{Smart Bird}, which is an efficient and effective Transformer approach with learnable sparse attention to automatically find and attend to important tokens. 
In \textit{Smart Bird}, we first use a single-head low-dimensional Transformer to learn a sketched attention matrix in an efficient way.
Next, we propose an attentive token sampling method to select important tokens to attend.
More specifically, we derive a sampling probability for each pair of tokens from the sketched attention matrix, where token pairs with higher attention weights have higher sampling probabilities.
We generate different sparse attention index matrices for different attention heads by randomly select tokens based on their sampling probabilities.
Finally, we collect token embeddings according to the index matrices to build the input embedding matrix of sparse attention networks.
In this way, the sparse attention network can better capture potentially important interactions between tokens to improve context understanding.
Extensive experiments conducted on six benchmark datasets for various tasks validate that \textit{Smart Bird} is both efficient and effective in text understanding.

The main contributions of this paper include:
\begin{itemize}
    \item We propose an efficient Transformer  with learnable sparse attention, which can capture the interactions between tokens that are informative for context modeling.
    \item We propose to learn sketched attention matrix with a low-dimensional Transformer to help recognize potentially important interactions between tokens.
    \item We propose an attentive token sampling method to build sparse attention index matrices according to the importance of token interactions for effective sparse attention.
    \item We conduct extensive experiments on six benchmark dataset and the results validate the efficiency and effectiveness of \textit{Smart Bird}.
\end{itemize}
\section{Related Work}

Since the vanilla Transformer is inefficient in processing long sequences, many methods explore to improve the efficiency of Transformer in different ways~\cite{tay2020efficient}.
One direction is computing a sparse attention matrix rather than a dense one by only computing attention on a sparse number of query and key vector pairs~\cite{child2019generating,beltagy2020longformer,zaheer2020big,zhang2021pooling}.
For example, Sparse Transformer~\cite{child2019generating} is a Transformer variant that combines local self-attention and stride attention.
It uses half of the attention heads to attend tokens within a region and the rest attention heads to attend tokens with certain strides.
Longformer~\cite{beltagy2020longformer} combines sliding window attention and global attention at certain positions to capture local and global contexts, respectively.
BigBird~\cite{zaheer2020big} further introduces a random attention mechanism that attends several  randomly selected pairs of tokens.
However, the token pairs that are randomly sampled or selected by fixed rules may not be helpful for context modeling, which limits the performance of these methods.

\begin{figure*}[!t]
  \centering 
      \includegraphics[width=0.8\linewidth]{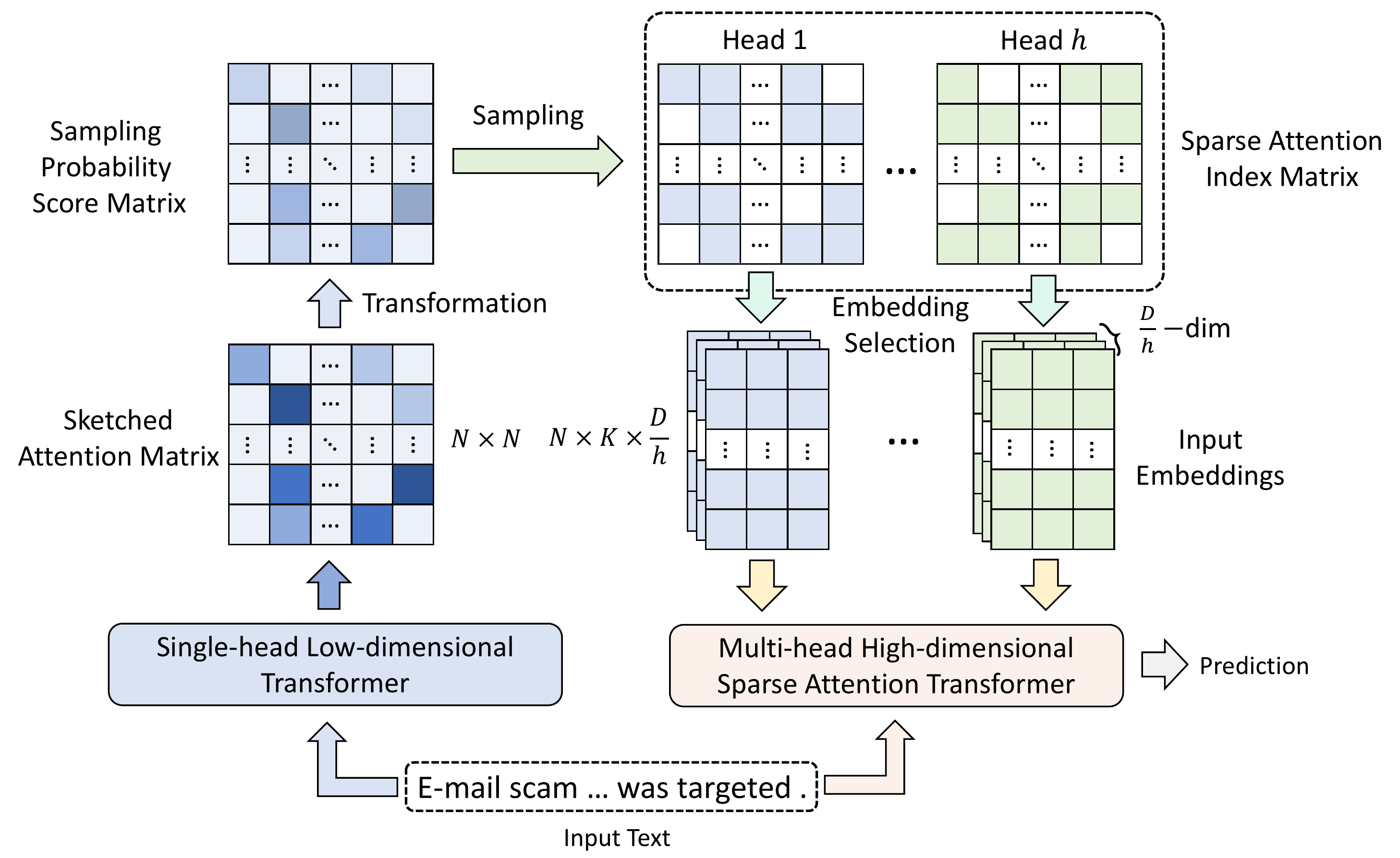}
  \caption{The overall framework of \textit{Smart Bird}.}\label{fig.model}

\end{figure*}

There are also many other ways to accelerate Transformers~\cite{Kitaev2020reformer,wang2020linformer}.
For example, Reformer~\cite{Kitaev2020reformer} uses hashing techniques to cluster input embeddings into different buckets based on their similarities, and then chunks the buckets using a certain length. The tokens only attend to same bucket in their own chunk and previous chunk.
Linformer~\cite{wang2020linformer} assumes that the self-attention matrix is low-rank, and it approximates the self-attention mechanism by using low-rank attention key and value projected by separate linear transformations.
Linear Transformer~\cite{katharopoulos2020transformers} uses kernel functions to approximate the self-attention mechanism.
It derives a kernel-based formulation of self-attention based on the matrix multiplication associative property and designs a simple kernel function to approximate the computation.
However,  these methods do not fully consider the characteristics of natural language and may be suboptimal in text understanding.
Different from existing efficient Transformers, \textit{Smart Bird} uses learnable sparse attention to capture important interactions between tokens, which can achieve both efficient and effective text modeling.

\section{Smart Bird}\label{sec:Model}

We introduce our proposed Transformer model named \textit{Smart Bird}, which can use learnable sparse attention to achieve both efficient and effective text understanding.
Its overall framework is shown in Fig.~\ref{fig.model}.
\textit{Smart Bird} first uses a low-dimensional single-head tiny Transformer to learn sketched self-attention matrix that indicates potentially important token interactions, and then uses an attentive token sampling method to build the sparse attention index matrices for different attention heads from the sketched self-attention matrix, and finally collects token embeddings according to the index matrices as the input of the multi-head sparse attention Transformer.
In this way, the sparse attention Transformer can attend to more informative tokens to better understand context information in an efficient way.
We then introduce the details of \textit{Smart Bird} in the following sections.

\subsection{Model Details}

The first step in \textit{Smart Bird} is computing the sketched self-attention matrix.
Since the computational complexity of Transformer is proportional to the hidden dimension, it is 
quite time-consuming to compute the self-attention matrix using the standard Transformer with high hidden dimensions.
Fortunately, we find that the self-attention matrix can be approximately computed by a very low-dimensional Transformer.
Thus, we first use a low-dimensional single-head Transformer to compute a sketched self-attention matrix\footnote{We do not use multi-head self-attention because it will increase the hidden dimension.}, which aims to find the potentially important interactions between tokens.
Assume the token sequence of an input text has $N$ tokens, which are denoted as $[w_1, w_2, ..., w_N]$.
The embedding sequence of these tokens is denoted as $\mathbf{E}=[\mathbf{e}_1, \mathbf{e}_2, ..., \mathbf{e}_N]$, where the dimension of embedding is $d$.
We use a $d$-dimensional Transformer to process this embedding sequence to obtain a hidden representation matrix $\mathbf{H}\in \mathbb{R}^{N\times d}$, which is formulated as 
\begin{equation}
    \mathbf{H}=\rm{Transformer}(\mathbf{W}^Q\mathbf{E},\mathbf{W}^K\mathbf{E},\mathbf{W}^V\mathbf{E}),
\end{equation}
where $\mathbf{W}^Q$, $\mathbf{W}^K$ and $\mathbf{W}^V$ are the query, key and value transformation matrices.
We further apply an attention pooling module to the hidden representation matrix $\mathbf{H}$ to learn a unified embedding $\mathbf{h}$ for the input text, and we use a linear transformation layer with softmax activation to predict the labels in the specific tasks for model training.\footnote{We assume the training task is a classification task here.}
By optimizing the loss functions of the training tasks, the tiny low-dimensional Transformer can learn an informative self-attention matrix $\mathbf{A}$ that indicates the importance of the interactions between all pairs of tokens.

The second step is building the sparse attention index matrix that indicates the tokens to be attended in the sparse attention mechanism.
An intuitive way is directly using the tokens with top attention weights in the sketched attention matrix.
However, different tokens usually also have different attention intensities, and each token may  attend to different numbers of tokens.
Thus, it may be suboptimal to simply select a certain number of tokens with top attention weights to attend. 
In addition, it is also challenging to generate multiple index matrices for different attention heads from a single attention matrix.
To solve these problems, we propose an attentive token sampling method to generate informative sparse attention index matrices for multiple attention heads.
We denote the sketched self-attention weight between the $i$-th and $j$-th tokens as $\alpha_{i,j}$.
The sampling probability score $p_{i,j}$ of this token pair is computed as follows:
\begin{equation}
    p_{i,j}=(\frac{1}{\log(\alpha_{i,j})})^2.
\end{equation} 
We use the logarithmic function to transform the raw attention weights because they have been normalized by the softmax function, and we use the squared inverse of logarithmic function to attend to more tokens with relatively higher attention weights.\footnote{We compare different sampling methods in experiments.}
For each token pair, we draw a sampling score $s_{i,j}$ from a uniform distribution as follows:
\begin{equation}
   s_{i,j}\sim U(0, p_{i,j}),
\end{equation} 
where $U(\cdot,\cdot)$ stands for the uniform distribution given a lower range and an upper range.
For each row in the sketched attention matrix, we choose the top $K$ tokens with the highest sampling scores to form the sparse attention index matrices.
The elements sparse attention index matrices are 1 if the corresponding token pair is selected and the rest are 0.
For different attention heads, we independently conduct the sampling process to generate different sparse attention index matrices, which can help more comprehensively capture the interactions between tokens by learning different attention patterns in different attention heads.

The final step aims to select token embeddings according to the sparse attention index matrices to form the input tensors of a multi-head sparse attention Transformer~\cite{zaheer2020big}.
Motivated by~\cite{zaheer2020big}, we use reshaped  inputs for efficient vectorized sparse attention computation.
More specifically, we select the embeddings of tokens that correspond to non-zero elements in the sparse attention index matrix to form the input embedding tensors.
For each attention head, the input tensor size is  $N\times K\times \frac{D}{h}$, where $D$ is the total hidden dimension and $h$ is the number of attention heads. 
If there are multiple Transformer layers, we apply the three steps described above to each layer.
We train the multi-head sparse attention Transformer in the target tasks, and we use it to generate the final predictions on the test samples.

\subsection{Complexity Analysis}

In this section, we provide some analysis on the theoretical computational complexity of \textit{Smart Bird}.
In \textit{Smart Bird}, the computational complexity of computing the sketched self-attention matrix is  $O(N^2\cdot d)$.\footnote{This step can be further accelerated by using low-rank self-attention, which will be explored in the future.}
The computational complexity of the sampling step is  $O(N^2)$, and the sparse attention Transformer has a complexity of $O(N\cdot K\cdot D)$.
The total computational cost of \textit{Smart Bird} is $O(N^2\cdot d+N\cdot K\cdot D)$.
Since the hidden dimension $d$ of the tiny Transformer is much smaller than the hidden dimension $D$ of a standard Transformer, the overall computational cost of \textit{Smart Bird} is much smaller than the vanilla Transformer with $O(N^2\cdot D)$ complexity, and is comparable with other efficient Transformer variants based on sparse attention mechanism if the input sequence length is not extremely long.

\section{Experiments}\label{sec:Experiments}

\subsection{Datasets and Experimental Settings}

We use  six benchmark  datasets for different tasks to conduct experiments.
The first one is AG's news\footnote{https://www.di.unipi.it/en/} (denoted as AG), which is a benchmark news topic classification dataset.
The second one is Amazon~\cite{he2016ups} (we use the ``Electronics'' domain), which is a widely used dataset for e-commerce review rating prediction.\footnote{https://jmcauley.ucsd.edu/data/amazon/}
The third one is IMDB~\cite{diao2014jointly}, which is a benchmark movie review rating prediction dataset.\footnote{https://github.com/nihalb/JMARS}
The fourth one is MIND~\cite{wu2020mind}\footnote{https://msnews.github.io/}, which is an English news dataset that contains news content and users' click behaviors.
We perform two tasks on this dataset, including news  topic classification based on news body and news recommendation task based on the relevance between clicked news and candidate news.
The fifth one is the CNN/DailyMail dataset~\cite{hermann2015teaching} (denoted as CNN/DM), which is a benchmark abstractive text summarization dataset. 
The sixth one is PubMed~\cite{cohan2018discourse}, which is a benchmark long document summarization dataset.
The statistics of the six datasets are shown in Table~\ref{dataset}.

\begin{table}[h]
\centering
\resizebox{1\linewidth}{!}{ 
\begin{tabular}{lccccc}
\Xhline{1.5pt}
\multicolumn{1}{c}{\textbf{Dataset}} & \textbf{\#Train} & \textbf{\#Val} & \textbf{\#Test} & \textbf{Avg. \#word} & \textbf{\#Class} \\ \hline
AG                                   & 108k             & 12k            & 7.6k            & 44                   & 4                \\
Amazon                               & 40.0k            & 5.0k           & 5.0k            & 133                  & 5                \\
IMDB                                 & 108.5k           & 13.6k          & 13.6k           & 386                  & 10               \\
MIND (class.)                 & 128.8k           & 16.1k          & 16.1k           & 505                  & 18               \\
MIND (rec.)                & 2.232m           & 376.5k         & 2.371m          & 12                   & -                \\
CNN/DM                               & 287.1k           & 13.4k          & 11.5k           & 781                  & -                \\
PubMed                               & 108.5k           & 13.6k          & 13.6k           & 3016                 & -                \\ \Xhline{1.5pt}
\end{tabular} 
}
\caption{Statistics of datasets.}\label{dataset}
\end{table}

In our experiments, we use Glove~\cite{pennington2014glove} embeddings to initialize the token embeddings of the high-dimensional Transformer, and we use principal component analysis (PCA) to align with the dimension of the tiny Transformer.
The hidden dimensions of the tiny Transformer and the standard sparse attention are 4 and 256, respectively.
We use 2 layers for all models in classification tasks and 4 layers for both encoder and decoder in summarization tasks.
In the news recommendation task, following many prior works~\cite{wu2020mind} we use news titles to model news content.
We use different Transformers to replace the word-level and news-level self-attention networks in NRMS~\cite{wu2019nrms}.
The text truncation length is 30 and the user clicked news sequence truncation length is 50.
On the AG dataset, the input truncation length is set to 256.
On the other datasets with longer text lengths, the truncation length of the vanilla Transformer and other efficient variants are 512 and 4096, respectively.\footnote{The maximum text length of Transformer is limited by the GPU memory.}
The sampling token number $K$ is set to 20.
We use Adam~\cite{kingma2014adam} as the model optimizer with a learning rate of 1e-4.
Since the classes in some datasets (e.g., Amazon) are imbalanced, We use both accuracy and macro-Fscore as the metrics.
On the news recommendation task, we use AUC, MRR, nDCG@5 and nDCG@10 to evaluate recommendation performance.
On the text summarization tasks, we use the ROUGE-1, ROUGE-2 and ROUGE-L scores (shorten as R-1, R-2 and R-L) as metrics.
We report the average scores of 5 independent experiments.

\begin{table*}[!t]
\resizebox{0.98\textwidth}{!}{
\begin{tabular}{lcccccccc}
\Xhline{1.5pt}
\multicolumn{1}{c}{\multirow{2}{*}{\textbf{Methods}}} & \multicolumn{2}{c}{\textbf{AG}}      & \multicolumn{2}{c}{\textbf{Amazon}}      & \multicolumn{2}{c}{\textbf{IMDB}} & \multicolumn{2}{c}{\textbf{MIND}} \\ \cline{2-9} 
\multicolumn{1}{c}{}                         & Accuracy       & Macro-F        & Accuracy    & Macro-F    & Accuracy    & Macro-F  & Accuracy    & Macro-F    \\ \hline
Transformer        & 93.13$\pm$0.12 & 93.10$\pm$0.13 & 65.40$\pm$0.31 & 42.45$\pm$0.34 & 52.11$\pm$0.41 & 42.70$\pm$0.42 & 81.01$\pm$0.18 & 61.42$\pm$0.19 \\
Sparse Transformer & 92.89$\pm$0.12 & 92.86$\pm$0.14 & 65.88$\pm$0.32 & 42.77$\pm$0.36 & 52.42$\pm$0.46 & 43.43$\pm$0.47 & 81.61$\pm$0.22 & 63.41$\pm$0.22 \\
Longformer         & 92.47$\pm$0.12 & 92.43$\pm$0.13 & 65.51$\pm$0.37 & 42.56$\pm$0.42 & 52.34$\pm$0.41 & 43.31$\pm$0.43 & 81.33$\pm$0.23 & 63.12$\pm$0.25 \\
Big Bird           & 92.88$\pm$0.12 & 92.85$\pm$0.11 & 66.10$\pm$0.39 & 43.04$\pm$0.41 & 52.95$\pm$0.48 & 43.87$\pm$0.49 & 82.10$\pm$0.21 & 63.64$\pm$0.23 \\ \hline
Smart Bird         & \textbf{93.42}$\pm$0.10 & \textbf{93.39}$\pm$0.11 & \textbf{66.45}$\pm$0.31 & \textbf{43.62}$\pm$0.34 & \textbf{53.75}$\pm$0.38 & \textbf{44.57}$\pm$0.40 & \textbf{82.60}$\pm$0.15 & \textbf{64.26}$\pm$0.6 \\  \Xhline{1.5pt} 
\end{tabular}
}
\caption{Results of topic and sentiment classification. } \label{table.performance2} 
\end{table*}

\begin{table}[!t]
 \resizebox{1.0\linewidth}{!}{
\begin{tabular}{lcccc}
\Xhline{1.5pt}
\textbf{Methods} & \textbf{AUC}   & \textbf{MRR}   & \multicolumn{1}{l}{\small{\textbf{nDCG@5}}} & \multicolumn{1}{l}{\small{\textbf{nDCG@10}}} \\ \hline
NRMS             & 68.24          & 33.33          & 36.37             & 42.26          \\
FIM             & 68.35          & 33.44          & 36.45             & 42.34          \\ \hline
Transformer       & 68.27          & 33.38          & 36.42             & 42.30 \\ 
Sparse Transformer      & 68.05          & 33.19          & 36.18             & 42.12 \\ 
Longformer       & 67.96          & 33.03          & 36.13             & 41.97 \\ 
BigBird      & 68.12          & 33.24          & 36.32             & 42.21 \\ 
\hline
Smart Bird    & \textbf{68.89}          & \textbf{34.06}          & \textbf{37.10}             & \textbf{43.21} \\
\Xhline{1.5pt}
\end{tabular}
}
\caption{Results of news recommendation. } \label{table.performance3} 
\end{table}

\begin{table}[!t]
\resizebox{0.98\linewidth}{!}{
\begin{tabular}{lcccccc}
\Xhline{1.5pt}
\multicolumn{1}{c}{\multirow{2}{*}{\textbf{Method}}} & \multicolumn{3}{c}{\textbf{CNN/DM}} & \multicolumn{3}{c}{\textbf{PubMed}} \\ \cline{2-7} 
\multicolumn{1}{c}{}                        & R-1       & R-2       & R-L       & R-1     & R-2     & R-L    \\ \hline
Transformer                                          & 38.51                     & 16.08                                               & 35.90                                               & 34.43                                               & 11.84                                               & 31.76                                               \\
Sparse Transformer                                   & 38.02                     & 15.50                                               & 35.10                                               & 37.19                                               & 14.63                                               & 33.85                                               \\
Longformer                                           & 37.99                     & 15.34                                               & 35.28                                               & 37.04                                               & 14.41                                               & 33.82                                               \\
BigBird                                              & 38.57                     & 15.80                                               & 35.86                                               & 37.77                                               & 15.16                                               & 34.53                                               \\ \hline 
Smart Bird                                  &   \textbf{39.22}        &  \textbf{16.96}         &   \textbf{37.04}        &   \textbf{38.83}      &  \textbf{16.01}       &  \textbf{35.49}      \\ \Xhline{1.5pt}
\end{tabular}
}
\caption{Results of text summarization.} \label{table.performance4} 
\end{table}

\begin{table}[t]
\centering
\begin{tabular}{lc}
\Xhline{1.5pt}
\multicolumn{1}{c}{\textbf{Method}} & \multicolumn{1}{c}{\textbf{Complexity}} \\ \hline
Transformer                         & $O(N^2\cdot D)$                          \\
Sparse Transformer                          & $O(N\sqrt{N} \cdot D)$                                             \\
Longformer                          & $O(N\cdot K \cdot D)$                                             \\
Big Bird                             & $O(N\cdot K \cdot D)$                                             \\
Smart Bird                      & $O(N^2\cdot d+N\cdot K \cdot D)$                         \\ \Xhline{1.5pt}
\end{tabular}
\caption{Complexity of different methods. $K$ is sentence length, $M$ is the number of sentences in a document, $T$ is the number of positions for sparse attention, and $d$ is the hidden dimension.}\label{complexity}
\end{table}

\subsection{Performance Evaluation}
We compare \textit{Smart Bird} with the vanilla  Transformer model and its several efficient variants based on sparse attention mechanism, including: 
(1) \textit{Sparse Transformer}~\cite{child2019generating}, a sparse attention based Transformer model that uses a combination of local attention and stride attention;
(2) \textit{Longformer}~\cite{beltagy2020longformer}, a Transformer variant based on sliding window attention and global attention at a few positions;
(3) \textit{Big Bird}~\cite{zaheer2020big}, a Transformer variant that integrates local attention, global attention and random attention mechanisms.
In addition, on the MIND datasets, we compare two additional SOTA news recommendation methods, i.e., NRMS~\cite{wu2019nrms} and FIM~\cite{wang2020fine}, to provide benchmarks for the comparison.
The performance of different methods on different datasets are compared in Tables~\ref{table.performance2},~\ref{table.performance3} and~\ref{table.performance4}.
We have several observations from the results.
First, the performance of efficient Transformer baselines including \textit{Sparse Transformer}, \textit{Longformer} and \textit{Big Bird} outperform the vanilla Transformer in long document modeling (e.g., classification tasks on Amazon, IMDB and MIND).
This is because the input text length of the vanilla Transformer is limited by the computing resources, and many useful contexts cannot be exploited when using a relatively short text truncation length.
Second, the baseline methods based on sparse attention are inferior to the vanilla Transformer in short sequence modeling  (i.e., classification on AG and recommendation on MIND).
This is because these methods cannot fully capture  the important interactions between different tokens.
Third, \textit{Smart Bird} can achieve better performance than other compared methods on all  datasets in different tasks.
This is because \textit{Smart Bird} incorporates learnable sparse attention to better capture token interactions that may be important for context modeling.
These results demonstrate the effectiveness and generality of  \textit{Smart Bird}.

Furthermore, we compare the theoretical computational complexity of different methods in Table~\ref{complexity}.
Since the dimension $d$ is much smaller than $D$, the overall complexity of \textit{Smart Bird} is much smaller than the vanilla Transformer, and is comparable with other sparse attention based methods if the sequence length is not extremely long.\footnote{\textit{Smart Bird} can support up to 4096 tokens on a GeForce GTX 1080 ti GPU and 16384 tokens on a Tesla V100 GPU.}
These results show that \textit{Smart Bird} is also efficient.

\begin{figure}[!t]
  \centering
  \subfigure[Training time.]{
    \includegraphics[width=0.45\textwidth]{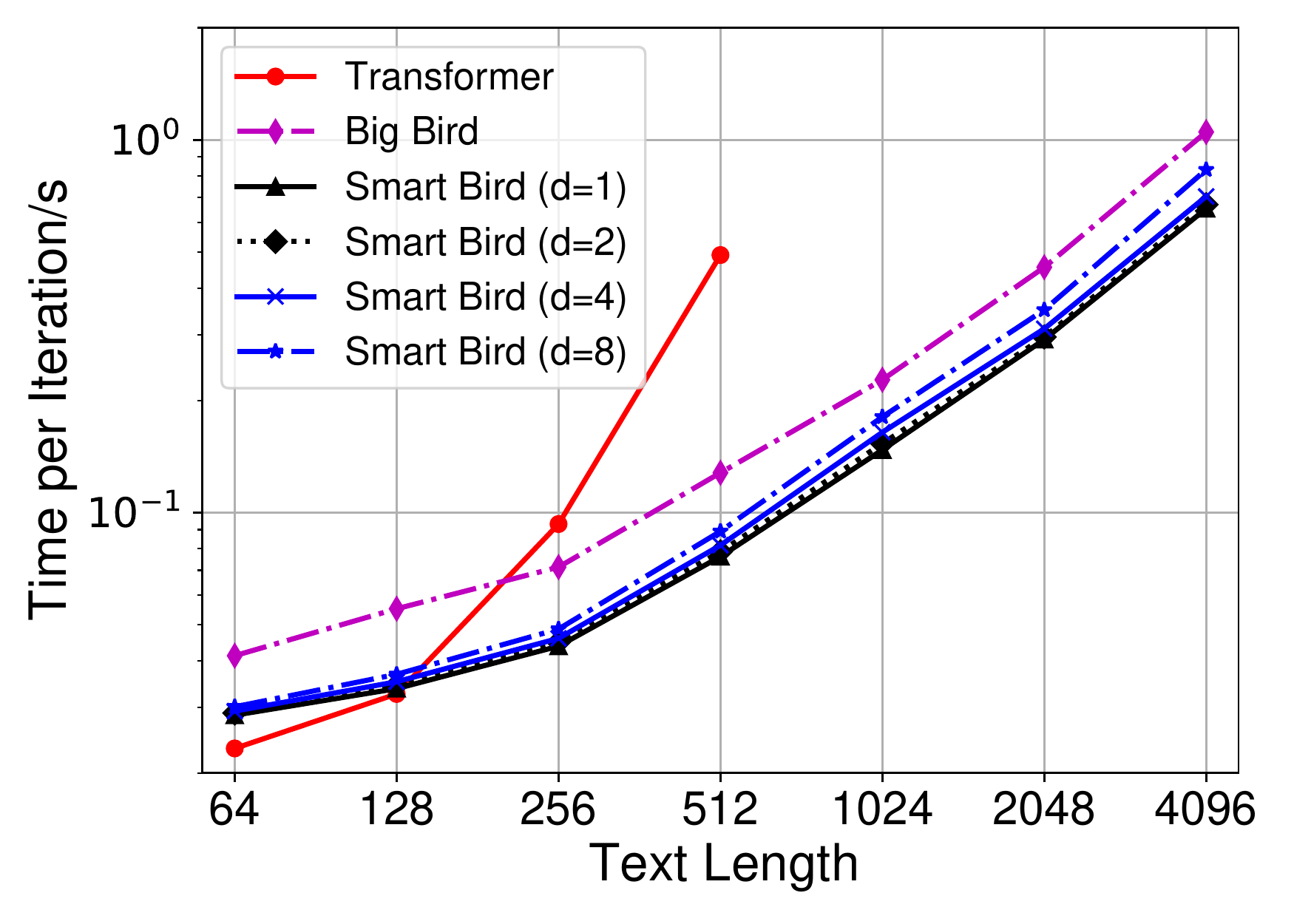}
  \label{fig.len1}
  }
   \subfigure[Model performance.]{
      \includegraphics[width=0.45\textwidth]{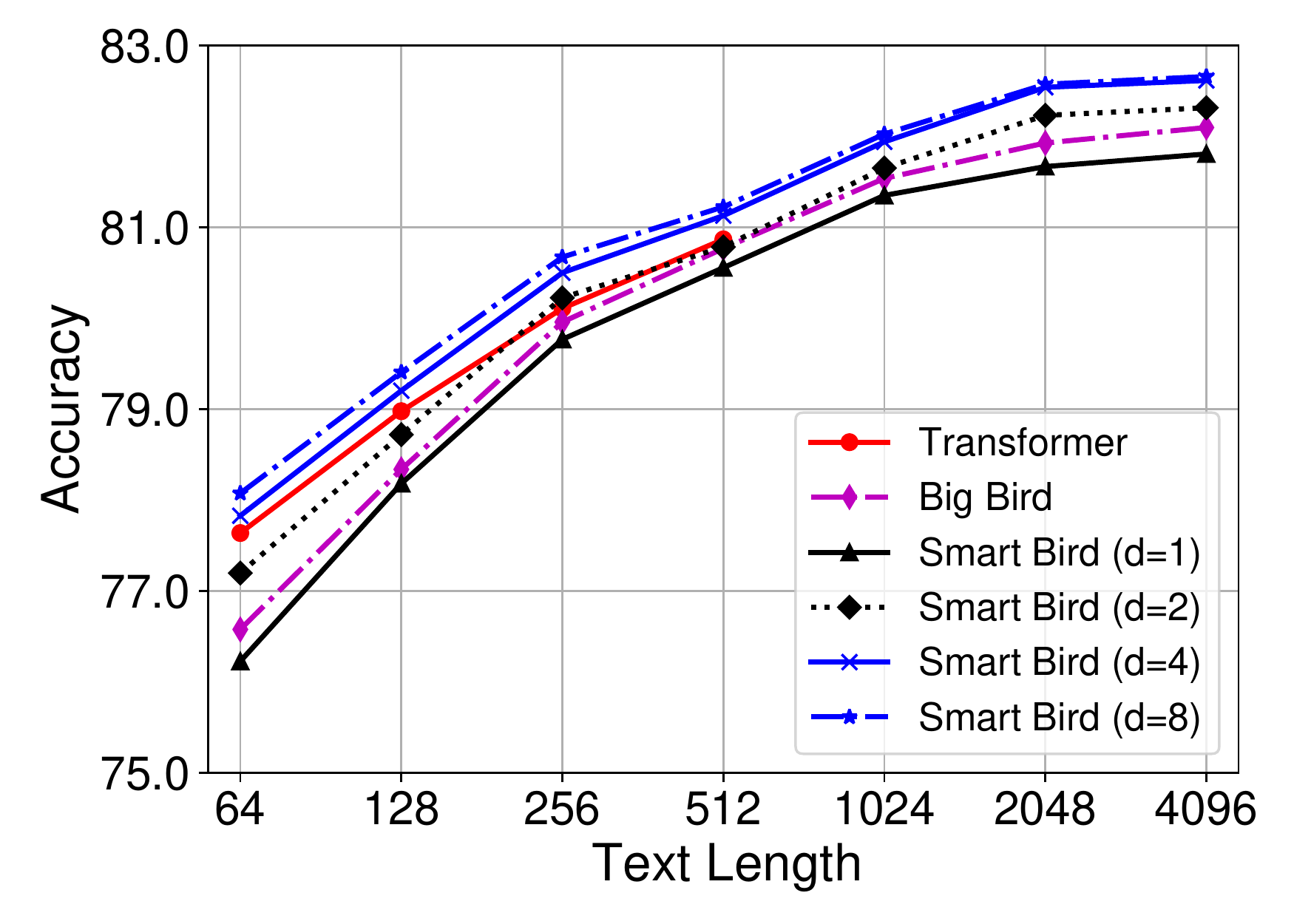}
  \label{fig.len2}
  }
  \caption{Training time and model performance under different text lengths. The training time axis (y-axis) uses logarithmic coordinates.}\label{fig.len}

\end{figure}

\subsection{Efficiency-Effectiveness Analysis}

Next, we present some more detailed analysis on the efficiency and effectiveness of \textit{Smart Bird}.
We compare the training time per iteration of each layer and the model performance with respect to different lengths of the input length.
We use the MIND dataset to conduct experiments since it has a relatively longer average text length.
The compared methods include Transformer, \textit{Big Bird} and \textit{Smart Bird} using different hidden dimensions in the  low-dimensional Transformer for learning the sketched self-attention matrix.
The results are shown in Fig.~\ref{fig.len}.
We observe that the performance of different methods consistently improves when longer texts are incorporated, which is because more context information can be modeled.
However, the training time of all methods also increases.
Fortunately, the total training time of \textit{Smart Bird} is smaller than Transformer and \textit{Big Bird} when the text length is longer than 256, which shows the efficiency of \textit{Smart Bird} in long text modeling.
In addition, we find that the performance of \textit{Smart Bird} increases when using higher hidden dimensions in the tiny Transformer.
This is because the sketched self-attention matrix may be inaccurate when tokens are only represented with very low-dimensional embeddings.
However, since the performance under $d=8$ is only marginally better than the performance under $d=4$, we prefer to choose $d=4$ for better efficiency.
Moreover, \textit{Smart Bird} consistently outperforms Transformer and \textit{Big Bird} under different text lengths.
It shows the effectiveness of \textit{Smart Bird} in modeling texts of different lengths.

\begin{figure}[!t]
  \centering
  \subfigure[Training time.]{
    \includegraphics[width=0.42\textwidth]{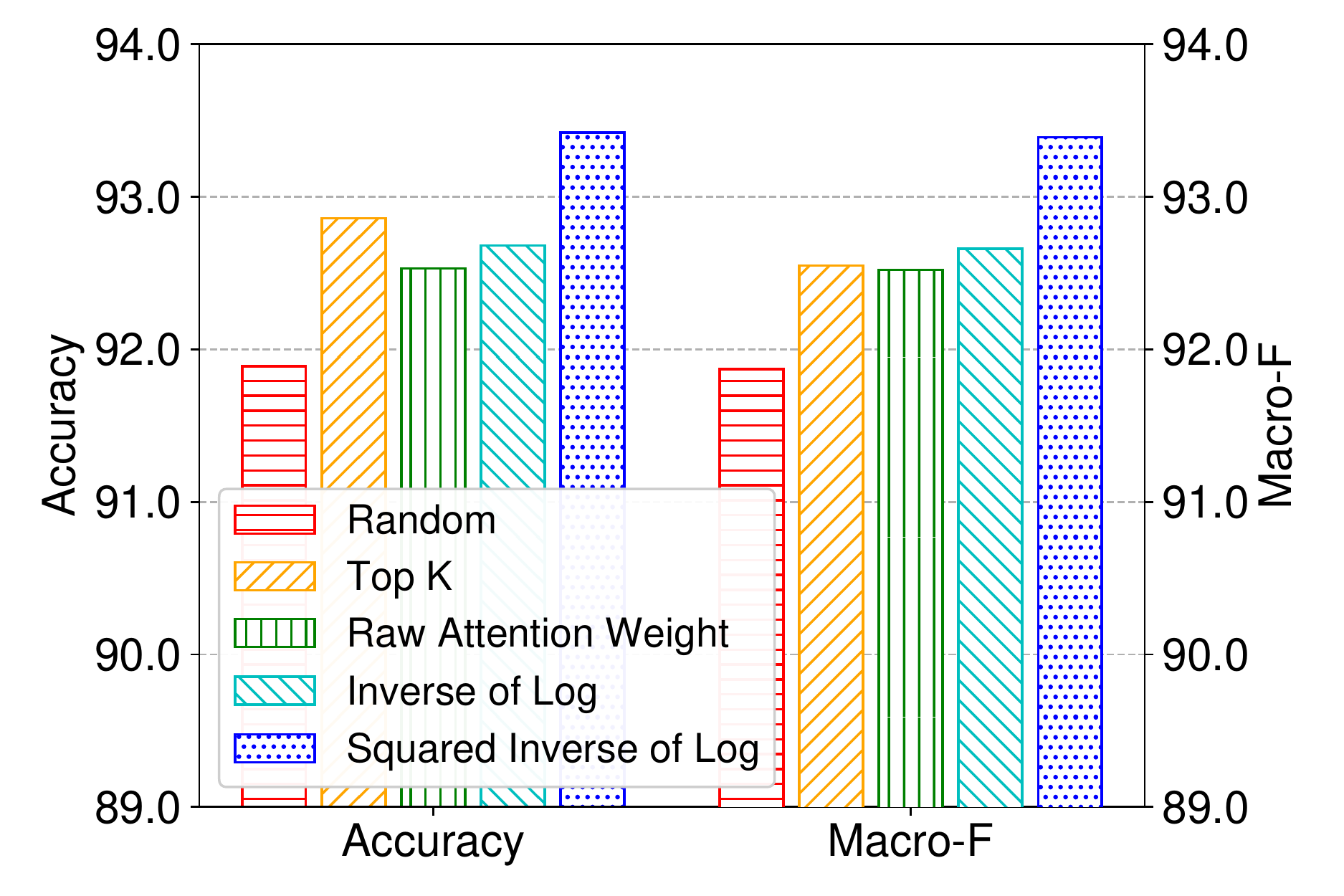}
  \label{fig.sample1}
  }
   \subfigure[Model performance.]{
      \includegraphics[width=0.42\textwidth]{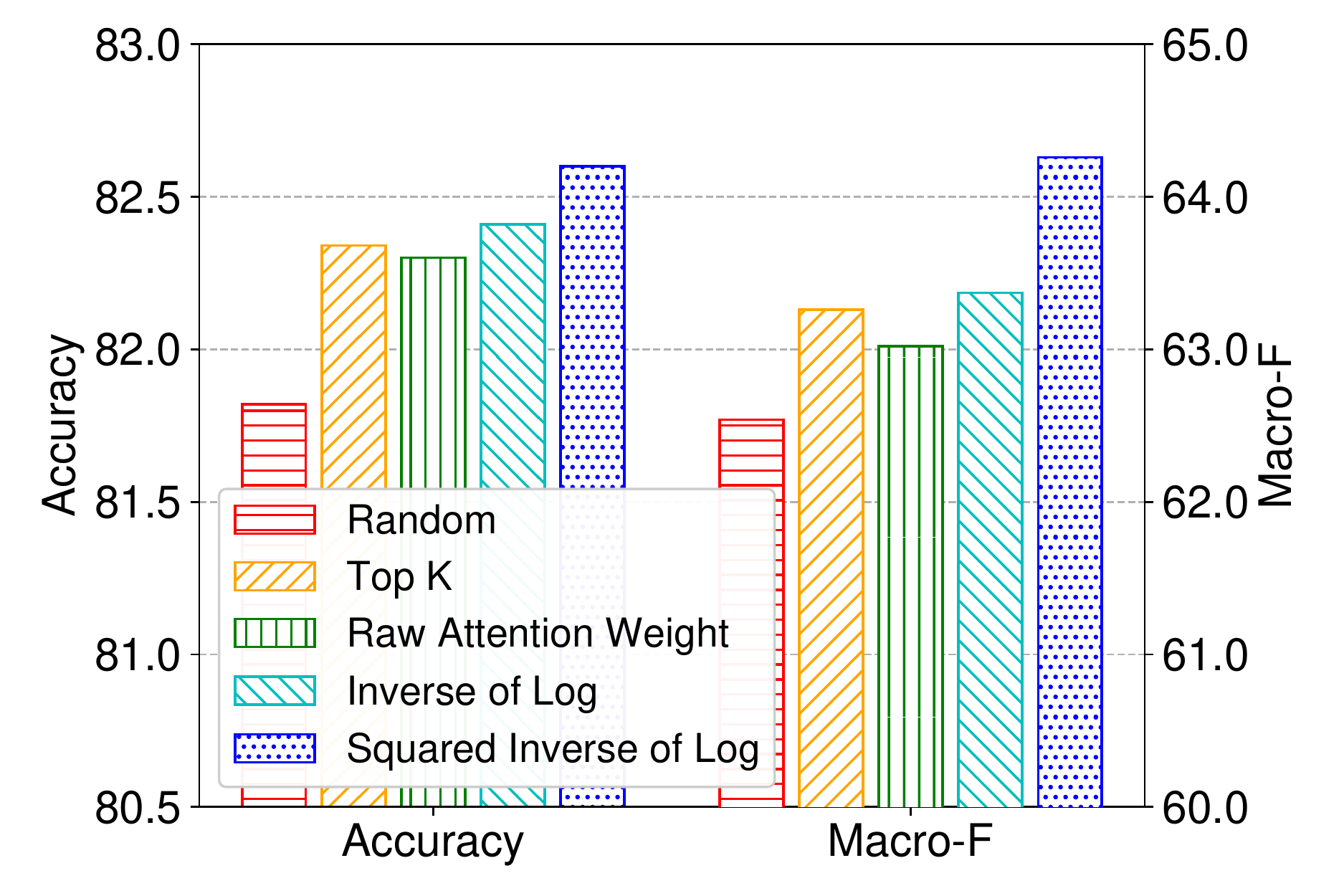}
  \label{fig.sample2}
  }

  \caption{Effectiveness of attentive token sampling.}\label{fig.sample}
\end{figure}

\begin{figure}[!t]
  \centering
  \subfigure[AG.]{
    \includegraphics[width=0.45\textwidth]{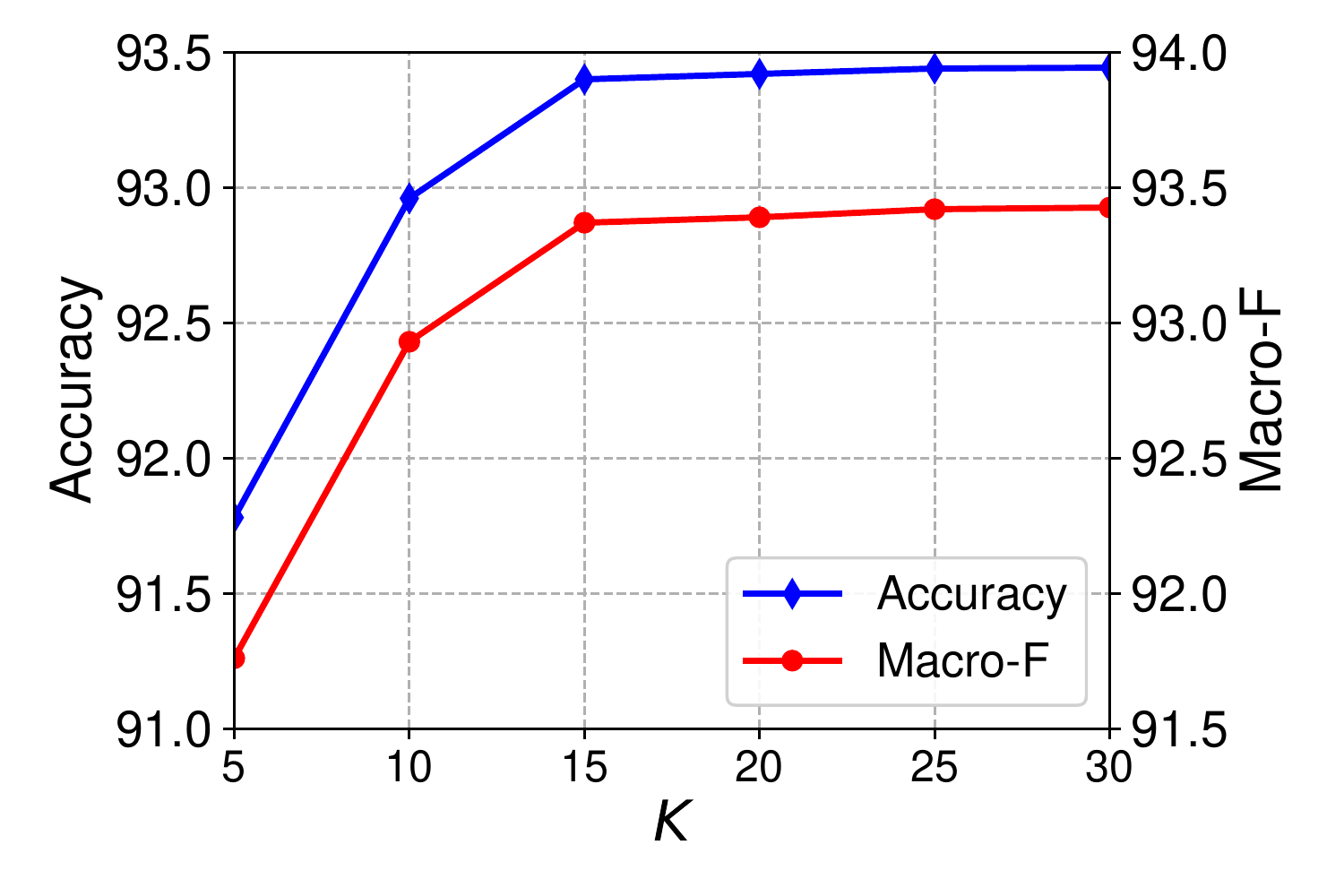}
  \label{fig.hyper1}
  }
   \subfigure[MIND.]{
      \includegraphics[width=0.45\textwidth]{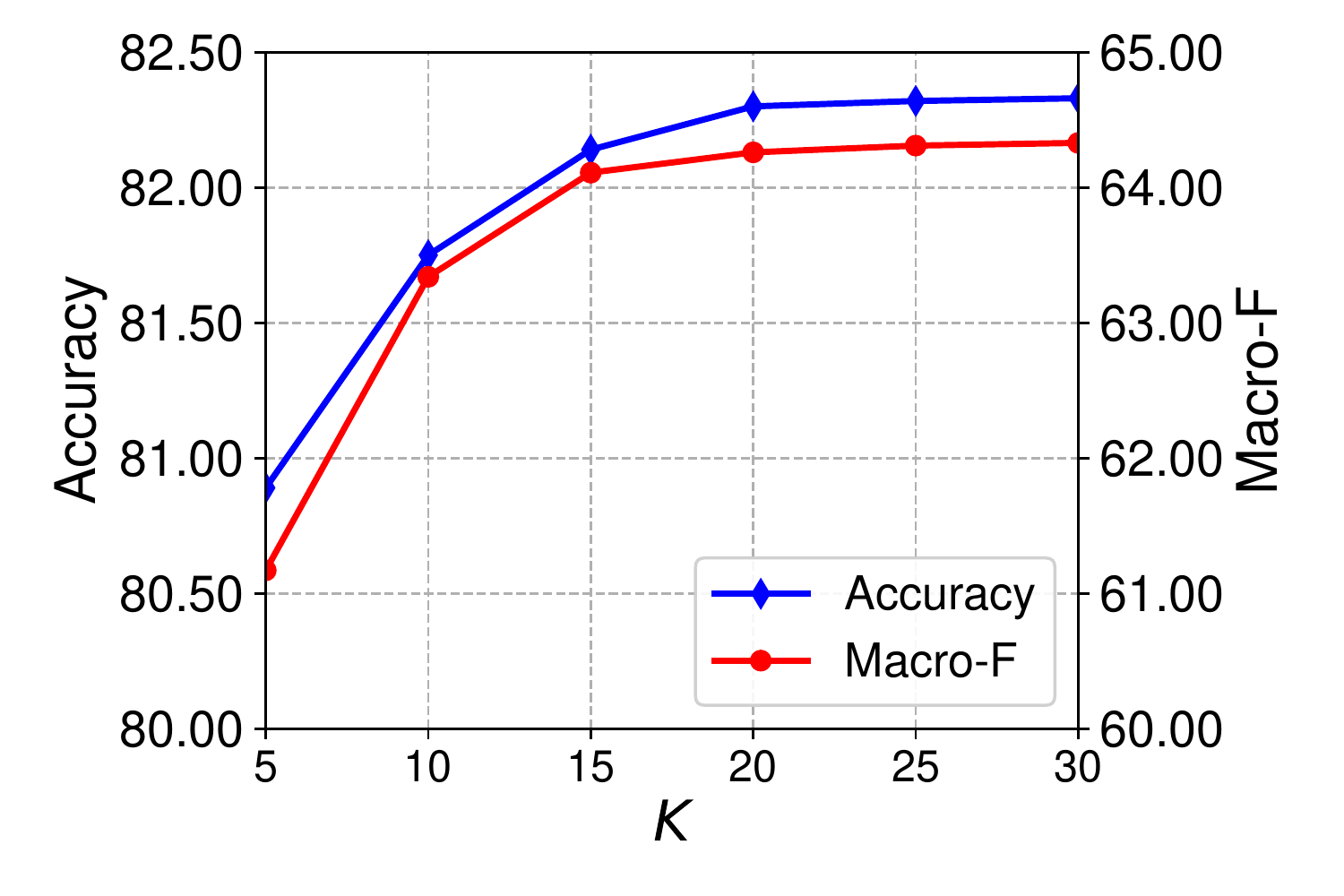}
  \label{fig.hyper2}
  }

  \caption{Influence of attending token number $K$.}\label{fig.hyper}
\end{figure}

\begin{figure*}[!t]
  \centering
  \subfigure[Raw sketched self-attention weights on AG.]{
    \includegraphics[width=0.4\textwidth]{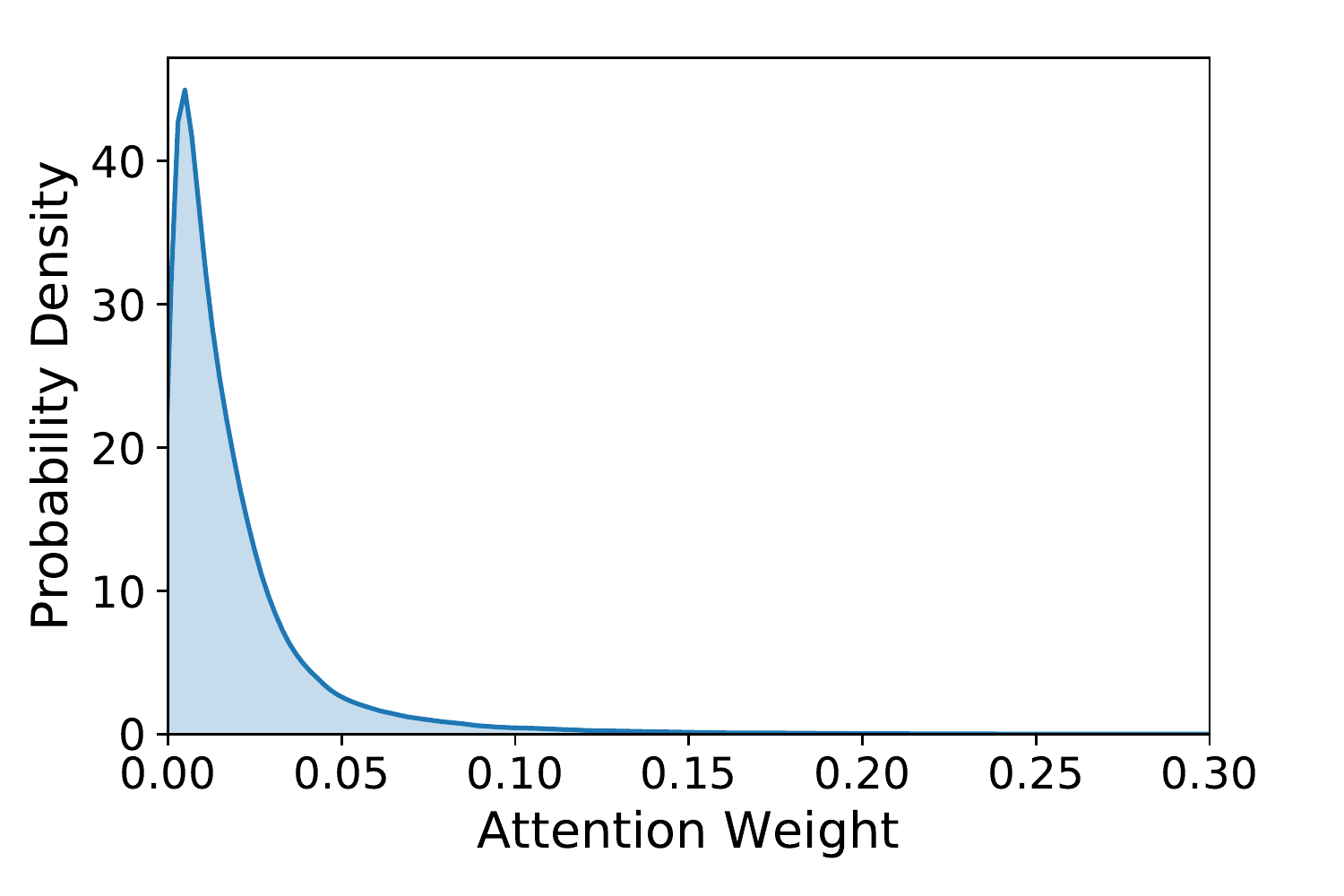}
  \label{fig.vi1}
  }
   \subfigure[Sampling scores on AG.]{
      \includegraphics[width=0.4\textwidth]{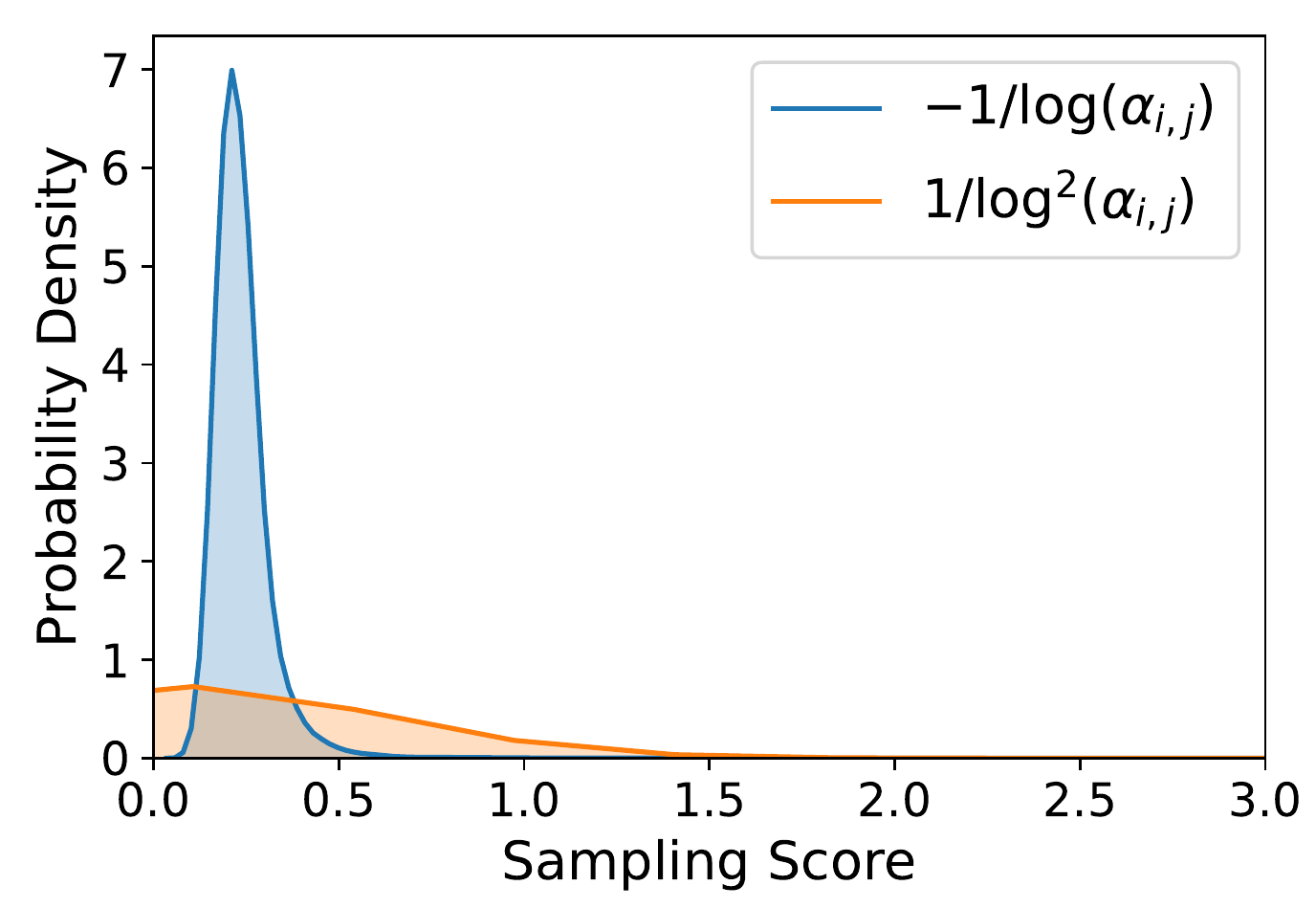}
  \label{fig.vi2}
  }
  \subfigure[Raw sketched self-attention weights on MIND.]{
    \includegraphics[width=0.4\textwidth]{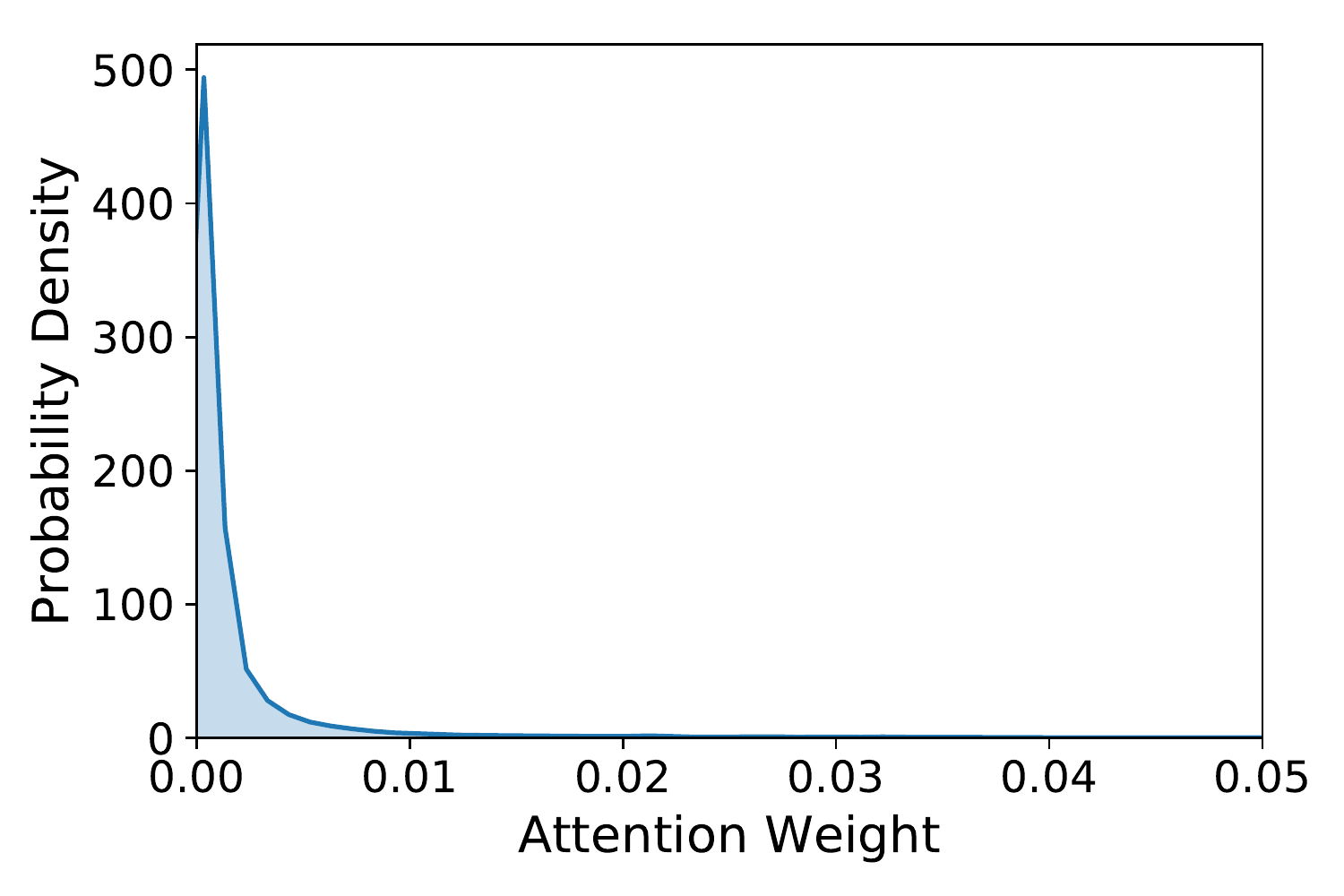}
  \label{fig.vi3}
  }
   \subfigure[Sampling scores on MIND.]{
      \includegraphics[width=0.4\textwidth]{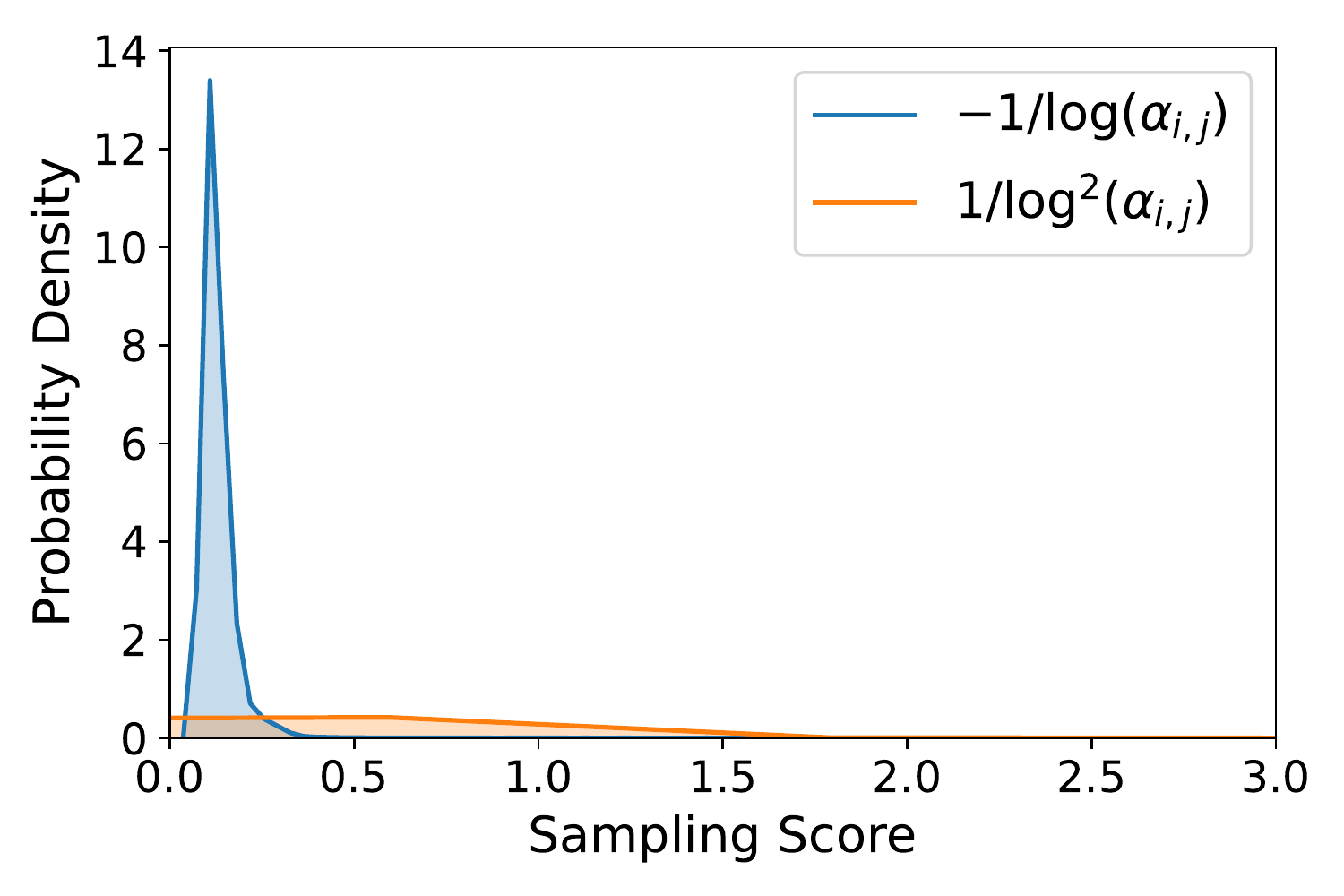}
  \label{fig.vi4}
  }

  \caption{Distributions of the raw self-attention weights and sampling scores using different computing functions on the AG and MIND datasets.}\label{fig.vi}
\end{figure*}

\subsection{Effect of Attentive Token Sampling}

We then verify the effectiveness of the attentive token sampling method in \textit{Smart Bird}.
We compare the performance of \textit{Smart Bird} with its variants using other token sampling strategies, including: (a) random, using randomly selected token pairs; (b) top K, using $K$ token pairs with the top attention weights in each row of the sketched attention matrix; (c) raw attention weight, using the raw attention weights as the sampling score in Eq. (3); (d) inverse of log, using the inverse of the logarithmic of attention in Eq.~(2); (e) squared inverse of log, the sampling method used in \textit{Smart Bird}.
The results on the AG and MIND datasets are shown in Fig.~\ref{fig.sample}.
We find that random attention yields the worst performance.
This is because randomly selected token pairs are usually less informative for context modeling.
In addition, using the raw attention weights is also suboptimal.
This is because the raw attention weights are normalized by softmax and very few tokens can gain high attention weights, which is not beneficial for comprehensively finding important interactions between tokens.
Besides, using the top K sampling strategy is also not optimal.
This is because the tiny Transformer may omit some useful contexts due to the dimension limit.
Thus, it may be better to explore more tokens pairs in sparse attention rather than exploit the top attention token pairs only.
Moreover, we find squared inverse of log function is better than the inverse of log function.
It may be because the latter one is too smooth and the squared version can help better capture important token interactions.

\subsection{Influence of Sampling Token Number}

We then study the influence of the number of sampling tokens in each row of self-attention matrix (i.e., $K$) on model performance.
We report the results of \textit{Smart Bird} under different values of $K$ in Fig.~\ref{fig.hyper}.
We find the model performance first improves rapidly when the value of $K$ increases.
This is because more informative interactions between tokens can be considered by \textit{Smart Bird} in text understanding.
However, when the number of $K$ is larger than 20, the performance gain is somewhat marginal.
Since the computational cost is proportional to $K$, we choose $K=20$ to balance model performance and computational cost.

\subsection{Weight Visualization}

Finally, we visualize the distributions of sketched self-attention weights as well as sampling scores in \textit{Smart Bird} to help better understand the attentive token sampling mechanism.
The raw sketched self-attention weights as well as the sampling scores computed by two different methods are shown in Fig.~\ref{fig.vi}.
We find that the distributions of raw attention weights are long-tail, and only a very small part of token pairs can gain high attention weights.
If we use the inverse of logarithmic function to compute sampling scores, their distributions are  centered insufficient to distinguish between important and unimportant token pairs.
On the contrary, using the squared version can help better discriminate the importance of token pairs, which can empower  the subsequent sampling process to build high-quality sparse attention index matrices.

\section{Conclusion and Future Work}\label{sec:Conclusion}

In this paper, we propose an efficient and effective Transformer variant named \textit{Smart Bird}, which can smartly attend to important token pairs based on a learnable sparse attention mechanism.
\textit{Smart Bird} first uses a low-dimensional tiny Transformer to compute a sketched self-attention matrix, and then uses an attentive token sampling method to select potentially important token pairs to be attended by a standard sparse attention Transformer.
\textit{Smart Bird} can effectively reduce the computational complexity of Transformer and can meanwhile recognize important interactions between tokens to help capture context information accurately.
Extensive experiments on six benchmark datasets for many different tasks fully validate the  efficiency and effectiveness of \textit{Smart Bird}.
In our future work, we plan to use low-rank techniques to accelerate the tiny Transformer for computing sketched self-attention weights and  improve the efficiency of \textit{Smart Bird} in processing extremely long sequences.

\bibliographystyle{acl_natbib}
\bibliography{acl2021}
 

\end{document}